\documentclass{article}
\PassOptionsToPackage{sort&compress, authoryear}{natbib}  

\usepackage[utf8]{inputenc}
\usepackage{microtype}

\usepackage{graphicx}
\graphicspath{{img/}}
\usepackage{subcaption}
\usepackage[x11names, dvipsnames]{xcolor}

\usepackage{chngcntr}

\newlength{\colsep}

\usepackage{xspace}
\makeatletter
\DeclareRobustCommand\onedot{\futurelet\@let@token\@onedot}
\def\@onedot{\ifx\@let@token.\else.\null\fi\xspace}

\def\ie{i.e\onedot}

\def\wrt{w.r.t\onedot}

\makeatother

\usepackage{amssymb}
\usepackage{amsmath}
\usepackage{dsfont}

\AtBeginDocument{%
  \abovedisplayskip 6pt plus 2pt minus 5pt%
  \belowdisplayskip \abovedisplayskip
  \abovedisplayshortskip  0pt plus 3pt%
  \belowdisplayshortskip  4pt plus 2pt minus 3pt%
}

\usepackage{multirow}
\usepackage{booktabs}
\usepackage{siunitx}
\usepackage{setspace}
\usepackage{etoolbox}

\usepackage{enumitem}

\usepackage{csvsimple}
\usepackage{ifthen}

\usepackage{adjustbox}

\usepackage{tikz}
\usetikzlibrary{%
  calc,
  shapes,
  positioning,
  fit,
  scopes,
  intersections,
  3d,
  plotmarks,
  external,
}

\usepackage{pgfplots}
\usepgfplotslibrary{%
  colorbrewer,
  external,
  fillbetween,
  groupplots,
}

\makeatletter
\tikzset{%
vector fill/.store in=\@vector@fill,
vector fill/.default=white,
vector hidden/.style={vector fill},
vector/.style={vector hidden/.append style={#1}},
pics/vector/.style args={#1/#2/#3/#4}{code={%
  \tikzset{vector hidden}%
  \node[fit={(0,0) (#1, #2)}, draw, inner sep=0pt, outer sep=0pt, fill=\@vector@fill] (\tikz@fig@name) at (0,0) {};%
  \notblank{#3}{\node[below=2pt of \tikz@fig@name] {#3};}{}%
  \notblank{#4}{\node[above=2pt of \tikz@fig@name] {#4};}{}%
}},
/layers/.cd,
fill/.store in=\@layer@fill,
fill/.default=white,
draw/.store in=\@layer@draw,
draw/.default=black,
width/.store in=\@layer@width,
width/.default=0.1,
height/.store in=\@layer@height,
height/.default=1.5,
depth/.store in=\@layer@depth,
depth/.default=1.0,
up label/.store in=\@layer@uplabel,
down label/.store in=\@layer@downlabel,
portrait/.store in=\@layer@portrait,
portrait/.default=,
layer hidden/.style={fill, draw, width, height, depth, portrait},
layer/.style={layer hidden/.append style={#1}},
/tikz/.cd,
pics/layer/.style={code={%
  \tikzset{/layers/.cd, layer hidden, #1}%
  \node[inner sep=0pt, outer sep=0pt] (\tikz@fig@name) {%
    \begin{tikzpicture}[transform shape]%
    \draw[fill=\@layer@fill, draw=\@layer@draw] (0,0,0) -- ++(\@layer@width, 0, 0) -- ++(0,\@layer@height,0) -- ++(-\@layer@width,0,0) -- cycle;
    \draw[fill=\@layer@fill, draw=\@layer@draw] (\@layer@width,0,0) -- ++(0, 0, -\@layer@depth) -- ++(0,\@layer@height,0) -- ++(0,0,\@layer@depth) -- cycle;
    \draw[fill=\@layer@fill, draw=\@layer@draw] (0,\@layer@height,0) -- ++(0, 0, -\@layer@depth) -- ++(\@layer@width,0,0) -- ++(0,0,\@layer@depth) -- cycle;
    \ifdefempty{\@layer@portrait}{}{%
      \begin{scope}[canvas is yz plane at x=\@layer@width]%
      \@layer@portrait%
      \end{scope}}%
    \end{tikzpicture}%
  };
  \notblank{\@layer@downlabel}{\node[below=-2pt of \tikz@fig@name] {\@layer@downlabel};}{}%
  \notblank{\@layer@uplabel}{\node[above=-2pt of \tikz@fig@name] {\@layer@uplabel};}{}%
}},
node distance=0.5cm,
>=latex,
semithick,
font=\footnotesize,
}
\makeatother

\pgfplotsset{compat=newest,%
  filter discard warning=false,%
  width=\linewidth,
  height=0.85\linewidth,%
  every axis plot post/.append style={thick},
  ymajorgrids,
  grid style={dashed}, 
  minimal plot grid/.style={
    y axis line style={opacity=0},
    axis x line*=bottom,
    x axis line style={black},
  },
  minimal plot grid,
  y tick label style={
    /pgf/number format/.cd,
    fixed,
    fixed zerofill,
    precision=1,
    /tikz/.cd,
    font=\scriptsize,
  },
  x tick label style={
    /pgf/number format/.cd,
    fixed,
    fixed zerofill,
    precision=1,
    /tikz/.cd,
    font=\scriptsize,
  },
  label style={font=\scriptsize},
  legend style={font=\scriptsize},
  /pgfplots/legend pos/south center/.style={/pgfplots/legend style={at={(0.5,0.03)},anchor=south}},
  /pgfplots/legend pos/north center/.style={/pgfplots/legend style={at={(0.5,0.97)},anchor=north}},
  /pgfplots/legend pos/outer north center/.style={/pgfplots/legend style={at={(0.5,1.05)},anchor=south}},
  /pgfplots/legend pos/outer east/.style={/pgfplots/legend style={at={(1.05,.5)},anchor=west}},
}

\newlength{\subfigwidth}
\newlength{\subfigheight}
\newsavebox{\imagebox}

\newcommand{\graphSurface}[3]{%
  \begin{tikzpicture}[every path/.style={line width=0.1pt}]%
  \csvreader[no head]{#1}{1=\a, 2=\b, 3=\c, 4=\d, 5=\e, 6=\f, 7=\color}{%
    \ifthenelse{\equal{\color}{a}}%
    {\def\col{black}%
      \draw[color=\col,opacity=.5] (\a, \b, \c) -- (\d, \e, \f) node {};%
    }%
    {\ifthenelse{\equal{\color}{b}}%
      {\def\col{red}}
      {\def\col{blue!60!black}}
      \draw[color=\col,opacity=1.] (\a, \b, \c) -- (\d, \e, \f) node {};%
    }%
  }%
  \csvreader[no head]{#2}{1=\g, 2=\h, 3=\i}{%
    \draw[draw=none, fill=blue!60!black] (\g, \h, \i) circle (#3pt);%
  }%
  \end{tikzpicture}%
}

\newcommand{\graphImg}[2]{%
  \begin{tikzpicture}[every path/.style={line width=2.5pt}]%
  \def\strick{.5}
  \def\accCol{white}
  \def\fnCol{Goldenrod}
  \def\fpCol{ProcessBlue}
  \selectcolormodel{cmyk}
  
  \csvreader[no head]{#1}{1=\i, 2=\j, 3=\r, 4=\g, 5=\b}{
    \def\color{rgb:red,\r;green,\g;blue,\b}
    \fill[opacity=1., fill=\color, draw=\color] (\i-\strick,\j-\strick) rectangle (\i+\strick,\j+\strick);
  }
  \csvreader[no head]{#2}{1=\o, 2=\p, 3=\q, 4=\r, 5=\color}{
    \ifthenelse{\equal{\color}{a}}%
    {
      \draw[opacity=1., color=\accCol] (\o, \p) -- (\q,\r);%
    }%
    {\ifthenelse{\equal{\color}{b}}%
      {\def\col{\fpCol}}
      {\def\col{\fnCol}}
      \draw[opacity=1., color=\col, line width=4pt, dashed] (\o, \p) -- (\q,\r);%
    }%
  }
  \csvreader[no head]{#1}{1=\i, 2=\j, 3=\r, 4=\g, 5=\b}{
    \fill[opacity=1., color=\accCol] (\i, \j) circle (3pt);%
  }
  \end{tikzpicture}%
}

\newcommand{\plotNodes}[1]{%
\begin{tikzpicture}
\begin{axis}[
  width=3.5cm,
  height=3.5cm,
  xlabel = $x'$,
  ylabel = $y'$,
  zlabel = $z'$,
  axis line style={black, opacity=1},
  axis lines=left,
  view/h = 30,
  view/v = 30,
  ticks = none,
  mark size=5,
]
\addplot3+[
  scatter,
  only marks,
  point meta=explicit symbolic,
  mark size=1pt,
  scatter/classes={
    a={mark=pentagon*, Dark2-A},
    b={mark=square*, Dark2-B},
    c={mark=triangle*, Dark2-C},
    d={mark=*, Dark2-D}
  },
] table[col sep=comma, meta index=3]{#1};
\end{axis}
\end{tikzpicture}
}

\newcommand{\graphCommunity}[3][]{%
  \begin{tikzpicture}[every path/.style={line width=0.2pt},#1]
  \csvreader[no head]{#2}{1=\o, 2=\p, 3=\q, 4=\r, 5=\color}{%
    \ifthenelse{\equal{\color}{a}}%
    {\def\col{black}%
      \draw[color=\col,opacity=.25] (\o, \p) -- (\q,\r) node {};%
    }%
    {\ifthenelse{\equal{\color}{b}}%
      {\def\col{red}}
      {\def\col{blue!60!black}}
      \draw[color=\col,opacity=.7] (\o, \p) -- (\q,\r) node {};
    }%
  }%
  \csvreader[no head]{#3}{1=\t, 2=\u, 3=\color}{%
    \ifthenelse{\equal{\color}{a}}%
    {\def\col{Dark2-A}\def\mark{pentagon*}}%
    {\ifthenelse{\equal{\color}{b}}%
      {\def\col{Dark2-B}\def\mark{square*}}%
      {\ifthenelse{\equal{\color}{c}}%
        {\def\col{Dark2-C}\def\mark{triangle*}}
        {\def\col{Dark2-D}\def\mark{*}}
      }
    }
    \node[mark size=.3pt,color=\col] at (\t, \u) {\pgfuseplotmark{\mark}};
  }%
  \end{tikzpicture}%
}

\usepackage{hyperref}

\usepackage[accepted]{icml2019}

\makeatletter
\patchcmd{\@makecaption}
{\vskip 10pt}
{\vskip 2.5pt plus 5pt}
{}{}
\makeatother

\icmltitlerunning{Graph Learning Network: A Structure Learning Algorithm}

\begin{document}

\twocolumn[%
\icmltitle{Graph Learning Network: A Structure Learning Algorithm}




\begin{icmlauthorlist}
\icmlauthor{Darwin Saire Pilco}{ic}
\icmlauthor{Adín Ramírez Rivera}{ic}
\end{icmlauthorlist}

\icmlaffiliation{ic}{Institute of Computing, University of Campinas, Campinas, Brazil}

\icmlcorrespondingauthor{Darwin Saire Pilco}{darwin.pilco@ic.unicamp.br}
\icmlcorrespondingauthor{Adín Ramírez~Rivera}{adin@ic.unicamp.br}

\icmlkeywords{Graph Prediction, Graph Structure Learning, Graph Neural Network}

\vskip 0.3in
]

{%
  \let\thefootnote\relax%
  \footnotetext{Code available at \url{https://gitlab.com/mipl/graph-learning-network}.}%
}

\printAffiliationsAndNotice{}  

\begin{abstract}
Recently, graph neural networks (GNNs) have proved to be suitable in tasks on unstructured data. Particularly in tasks as community detection, node classification, and link prediction.
However, most GNN models still operate with static relationships.
We propose the Graph Learning Network (GLN), a simple yet effective process to learn node embeddings and structure prediction functions. 
Our model uses graph convolutions to propose expected node features, and predict the best structure based on them. 
We repeat these steps recursively to enhance the prediction and the embeddings.
\end{abstract}

\section{Introduction}
\label{sec:intro}
When working on unstructured information, commonly, graphs are employed because they can represent this information naturally.
For instance, in social networks, system recommendations, and link prediction, graphs can capture the relationship between entities.
In order to work on this type of information, deep models on graphs were created \citep{Gori2005, Scarselli2009, Defferrard2016, Kipf2017}. 
These models take into account the information of each node and its neighborhood relationships when extracting new information (\ie, node embedding).
Unlike traditional models on graphs, which still work on a static domain (\ie, graphs without variation in the structure), \citet{Li2016, Marcheggiani2017, Ying2018, Bresson2018} began to work on dynamic domains (\ie, variable graphs). 
However, they still do not support extreme variations; \ie, complete changes in the structure of graphs in each layer.

\textbf{Related work.}
We classify the graph representation learning methods into two groups: \textit{generative models} that learn the graph relationship distribution from latent variables, and \textit{discriminative models} that predict the edge probability between pairs of vertices.

For generative models, the Variational Autoencoder (VAE) \citep{Kingma2014, Sohn2015} proved to be competent at generating graphs.
Thus, methods based on VAEs \citep{Kusner2017, Grover2018, Simonovsky2018, Bojchevski2018, DeCao2018, Kearnes2019} learn some probability distribution that fits and models the graph's relationships.
Other methods \citep{Li2018, You2018} propose auto-regressive models (\ie, generate node-to-node graphs) to generate graphs with a similar structure.
Nevertheless, we consider relevant to contrast ourselves with the generative methods since they aim to learn the structures (regardless of the difference in the final task).

In contrast to the first group, the discriminative models do not use conditional distributions to generate edges but directly aim to learn a classifier for predicting the presence of edges.
For this, a diversity of models based on GNNs \citep{Gori2005, Scarselli2009} were explored \citep{Battaglia2018, dgl.ai}.
For example, methods for recommendation systems on bipartite graphs were proposed by \citet{Berg2018}.  
\citet{Schlichtkrull2018} merged auto-encoder and factorization methods (\ie, use of scoring function) to predict labeled edges.  
Besides, diverse approaches try to take advantage of recurrent neural networks \citep{Monti2017}, and heuristic methods \citep{Zhang2018a, Donnat2018}.  
Different from previous methods, message-passing approaches \citep{Gilmer2017, Kipf2018, Battaglia2018} add edge embedding for each relationship between two nodes.
Similarly, we predict the edges of the graph based on an initial set of nodes and a configuration.  
However, we learn local and global transformations around the nodes, while transforming the features too, in turn, enhance the structure prediction. 

In this paper, we predict new structures from the local and global node embedding in the graph through a recurrent set of operations. 
In each application of our block, we adjust the graph's structure and nodes' features. 
In other words, we work with variable graphs to predict new structures.

\textbf{Contributions.}
(i)~Two prediction functions (for nodes' features and adjacency) that lets us extract the most probable structure given a set of points and their feature embeddings, respectively. 
(ii)~A recurrent architecture that defines our iterative process and our prediction functions. 
(iii)~An end-to-end learning framework for predicting graphs' structure given a family of graphs.
(iv)~Additionally, we introduce a synthetic dataset, \ie, 3D surface functions, that contains patterns that can be controlled and mapped into graphs to evaluate the robustness of existing methods. 

\section{Graph Learning Network}
\label{sec:gln}

Given a set of vertices $V = \{v_i \}$, such that every element $v_i$ is a feature vector, we intend to predict its structure as a set of edges between the vertices, $E = \left\{ \left(v_i, v_j \right) : v_i, v_j \in V \right\}$.  
In other words, we want to learn the edges of the graph $G = (V, E)$ that maximize the relations between the vertices given some prior patterns, \ie, a family of graphs. 

\begin{figure}[tb]
  \centering
  \resizebox{.8\linewidth}{!}{\begin{tikzpicture}[
  /layers/.cd,
  layer={height=1},
  /tikz/.cd,
  connect/.style={
    ->,
    rounded corners,
    shorten >=2pt,
    shorten <=2pt,
  }
]
\pgfdeclarelayer{bg}    
\pgfsetlayers{bg,main} 

\def\maxL{1}
\def\maxcellL{1}

\def\layerDist{1.25cm}
\def\streamDist{2cm}

\colorlet{fillcolor}{DodgerBlue1!50}
\colorlet{backgroundcolor}{gray!25}
\colorlet{graphcolor}{blue!50!gray}

\csdef{g0}{%
  \tikzset{every node/.style={circle, fill=graphcolor, inner sep=1.5pt}, every path/.style={draw=graphcolor}, node distance=5pt, shorten >=0pt, shorten <=0pt}%
  \node (a) at (.3,-.5) {};%
  \node (b) at (.5,-.75) {};%
  \node (c) at (.6,-.5) {};%
  \node (d) at (.1,-.3) {};%
  \node (e) at (.8,-.8) {};%
  \node (f) at (.8,-.2) {};%
}
\csdef{g1}{%
  \csuse{g0}%
  \draw (a) -- (b);%
}
\csdef{g2}{%
  \csuse{g1}%
  \draw (b) -- (c);%
}
\csdef{g3}{%
  \csuse{g2}%
  \draw (a) -- (c);%
}
\csdef{g4}{%
  \csuse{g3}%
  \draw (a) -- (d);%
  \draw (e) -- (f);%
}

\draw pic (a1) at (0,0) {layer={down label=$A^{(l)}$, up label={\scriptsize$n \times n$}, portrait={\csuse{g1}}}};

\draw pic[below=\streamDist of a1] (h1) {layer={down label=$H^{(l)}$, up label={\scriptsize$n \times d_l$}, fill=fillcolor}};

\node[right=\layerDist of h1] (h3) {$\vdots$};
\draw pic[above=.5\streamDist of h3] (h4) {layer={down label=$H^{(l)}_{1}$, up label={\scriptsize$n \times d_l$}, fill=fillcolor}};
\draw pic[below=.5\streamDist of h3] (h5) {layer={down label=$H^{(l)}_{k}$, up label={\scriptsize$n \times d_l$}, fill=fillcolor}};

\draw pic[right=\layerDist of h3] (h6) {layer={down label=$H^{(l)}_{\text{int}}$, up label={\scriptsize$n \times d_{l+1}$}, fill=fillcolor}};
\draw pic[right=\layerDist of h6] (h7) {layer={down label=$H^{(l)}_{\text{local}}$, up label={\scriptsize$n \times d_{l+1}$}, fill=fillcolor}};
\draw pic[] (h8) at (a1 -| h7) {layer={down label=$H^{(l)}_{\text{global}}$, up label={\scriptsize$n \times d_{l+1}$}, fill=fillcolor}};

\draw pic[right=\layerDist of h8] (a2) {layer={down label=$A^{(l+1)}$, up label={\scriptsize$n \times n$}, portrait={\csuse{g3}}}};

\draw pic[] (h9) at (h7 -| a2) {layer={down label=$H^{(l+1)}$, up label={\scriptsize$n \times d_{l+1}$}, fill=fillcolor}};

\draw[connect] (h1.east) -- (h3.west);
\draw[connect] (a1.east) -| ($(h1)!.45!(h3)$);

\coordinate (h13) at ($(h1)!.55!(h3)$);
\draw[connect, shorten <=0pt] (h13) |- (h4.west);
\draw[connect, shorten <=0pt] (h13) |- (h5.west);

\coordinate (h36) at ($(h3)!.45!(h6)$);
\node[draw, circle, minimum size=0.3cm, fill=white, inner sep=.5pt] (plus op) at (h36) {$+$};
\draw[connect] (h4.east) -| (plus op);
\draw[connect] (h3.east) -- (plus op);
\draw[connect] (h5.east) -| (plus op);

\draw[connect] (a1.east) -| ($(h6)!.5!(h7)$);

\draw[connect] (plus op) -- node[below] {$\eta_{l}$} (h6.west);

\draw[connect] (h6.east) -- node[below] {$\lambda_l$} (h7.west);
\draw[connect] (h7.east) -- (h9.west);
\draw[connect, shorten <=8pt, shorten >=17pt
] (h7.north) -- node[right] {$\gamma_{l}$} (h8.south);

\draw[connect] (h8.east) -- node[above] {$\rho_{l}$} (a2.west);
\draw[connect, shorten <=0pt] (h7 -| {$(h8)!.5!(a2)$}) -- ($(h8)!.5!(a2)$);

\begin{pgfonlayer}{bg}
  \coordinate (a) at ($(a1.north east)+(10pt, 10pt)$);
  \coordinate (b) at ({$(h5.south)+(0, -15pt)$} -| {$(a2.north west)+(-10pt,0pt)$});
  \draw[backgroundcolor, fill=backgroundcolor!50, thick, dashed, rounded corners] (a) rectangle (b);
  \node[anchor=south east] at (b) {Recurrent Block};
\end{pgfonlayer}

\end{tikzpicture}}
  \caption{%
    Our proposed method is a recurrent block. 
    We create a set of node embeddings $\big\{ H^{(l)}_i \big\}_{i=1}^k$ that are later combined to produce an intermediary representation $H^{(l)}_{\text{int}}$.
    Then, we use the updated node information with the adjacency information to produce a local embedding of the nodes information $H^{(l)}_{\text{local}}$ that is also the output $H^{(l+1)}$. 
    We also broadcast the information of the local embedding to produce a global embedding $H^{(l)}_{\text{global}}$.  
    We combine the local and global embeddings to predict the next layer adjacency $A^{(l+1)}$.
  }
  \label{fig:model}
\end{figure}
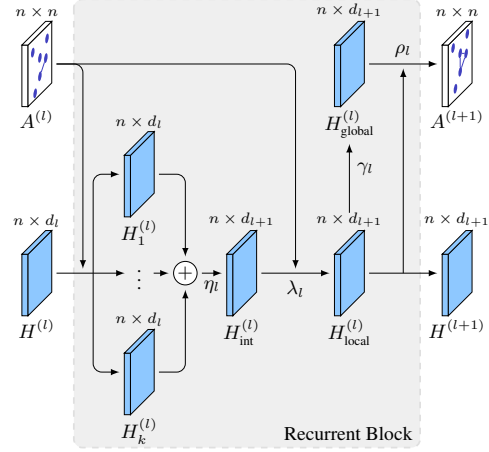

To achieve this, we perform two alternating tasks for a given number of times, akin to an expectation-maximization process.  
At each step, we transform the nodes' features through convolutions on the graph \citep{Kipf2017} using multi-kernels to learn better representations to predict their structure.  
Then, we merge the multiple node embedding and apply function transform \citep{Bai2019} on them that combines the local and global contexts for the embeddings.
Next, we use these transformed features (local and global) in a pairwise node method to predict the next structure, which is represented through an adjacency matrix.  
The learned convolutions on the graph represent a set of responses on the nodes that will reveal their relations.  
These responses are combined to create or delete connections between the nodes, and encoded into the adjacency matrix.  
The sequential application of these steps recover effective relations on nodes, even when trained on families of graphs.  
We represent this process in Fig.~\ref{fig:model}. 

\textbf{Node Embeddings.} 
\label{sec:node_embedding} 
At a given step~$l$ on the alternating process, we have $d_l$ hidden features, $H^{(l)} \in \mathbb{R}^{n \times d_l}$, for each of our $n$~nodes, and the set of edges (structure) encoded into an adjacency matrix $A^{(l)} \in [0,1]^{n \times n}$ that represents our graph.  
As introduced, our first step is to produce the features of the next step, $H^{(l+1)}$, through the embedding
\begin{equation}
  \label{eq:h-output}
  H^{(l+1)} = \lambda_l \left( \eta_l \left( H^{(l)}, A^{(l)} \right), A^{(l)} \right).
\end{equation}
Our embedding comprises to steps: extracting $k$ features for the nodes, and combining them into an intermediary embedding~(\ref{eq:inter}); and creating a local representation~(\ref{eq:local}). 
For the first step, we use convolutional graph operations \citep{Kipf2017}
\begin{equation}
  \label{eq:inter}
  H^{(l)}_{\text{int}} = \eta_l \left( H^{(l)}, A^{(l)} \right) = \sum_{i=1}^{k} \sigma_l \left( \tau \left( A^{(l)} \right) H^{(l)} W^{(l)}_i \right),
\end{equation}
where $k$ is the number of kernels, $W^{(l)}_i \in \mathbb{R}^{d_l \times d_{l+1}}$ is the learnable weight matrix for the $i$th convolutional kernel at the $l$th step, $\sigma_l(\cdot)$ is a non-linear function, and $\tau(\cdot)$ is a symmetric normalization transformation of the adjacency matrix, defined by
\begin{equation}
  \label{eq:sym-trans}
  \tau \left( {A}^{(l)} \right) = \left( \hat{D}^{(l)} \right)^{-\frac{1}{2}} \left( A^{(l)} + I_n \right) \left( \hat{D}^{(l)} \right)^{-\frac{1}{2}},
\end{equation}
where $\hat{D}^{(l)}$ is the degree matrix of the graph plus identity, that is,
\begin{equation}
  \hat{D}^{(l)} = D^{(l)} + I_n,
\end{equation}
where $D^{(l)}$ is the degree matrix of $A^{(l)}$, and $I_n$ is the identity matrix of size $n \times n$. 
Unlike previous work \citep{Kipf2017}, we are computing convolutions that will have different neighborhoods at each step defined by the changing $A^{(l)}$, in addition to multiple learnable kernels per layer.
In summary, this step allows us to learn a response function, defined by the weights $W^{(l)}_i$ of the $i$th kernel, that embed the node's features into a suitable form to predict the structure of the graph. 

The second step corresponds to create a local-context embedding from the intermediary representation~(\ref{eq:inter}) that depends on the current adjacency.  
We define our local context $\lambda_l$ as
\begin{equation}
\label{eq:local}
H^{(l)}_{\text{local}} = \lambda_l \left( H^{(l)}_{\text{int}}, A^{(l)} \right) = \sigma_l \left( \tau \left( A^{(l)} \right) H^{(l)}_{\text{int}} U^{(l)} \right),
\end{equation}
where $U^{(l)} \in \mathbb{R}^{d_{l+1} \times d_{l+1}}$ is the learnable weight matrix for the linear combinations of the nodes' features $H^{(l)}_{\text{int}}$.

\textbf{Adjacency Matrix Prediction.}
\label{sec:adj_pred}
After obtaining the nodes embeddings, $H^{(l)}_\text{local}$~(\ref{eq:local}), we use them to predict the next adjacency matrix~$A^{(l+1)}$ through
\begin{equation}
  \label{eq:graph}
  A^{(l+1)} = \rho_l \left( H^{(l)}_{\text{local}} \right) = \sigma_l \left( M^{(l)} \alpha_l \left( H^{(l)}_{\text{local}} \right) {M^{(l)}}^\top \right),
\end{equation}
where $M^{(l)} \in \mathbb{R}^{n \times n}$ is the weight matrix that produces a symmetric adjacency, $\alpha_l$ is a transformation that mixes global and local information within the graph, and $\cdot^\top$ denotes the transposition operator.

We broadcast the local information to all the nodes by assuming that all the nodes are connected, \ie, the adjacency on the graph would be $A^{(l)} = \mathds{1}$, and then using a convolution operation. We define the global context as
\begin{equation}
  \label{eq:global}
  H^{(l)}_{\text{global}} = \gamma_l \left( H^{(l)}_{\text{local}} \right) = \sigma_l \left( H^{(l)}_{\text{local}} Z^{(l)} \right),
\end{equation}
where $Z^{(l)} \in \mathbb{R}^{d_{l+1} \times d_{l+1}}$ is the learnable weight matrix.
This operation is similar to attention mechanisms previously used \citep{Bai2019}, yet, we use it as a broadcasting mechanism instead.

Finally, we merge both local~(\ref{eq:local}) and global~(\ref{eq:global}) contexts using a transformation function
\begin{equation}
  \label{eq:a_l}
  \alpha_l \left( H^{(l)}_{\text{local}} \right) = H^{(l)}_{\text{local}} Q^{(l)} \gamma_l \left( H^{(l)}_{\text{local}} \right)^\top,
\end{equation}
where $Q^{(l)} \in \mathbb{R}^{d_{l+1} \times d_{l+1}}$ is the learnable weight matrix. 
The intuition is that nodes similar to the global and local context should receive higher attention weights for the projection of a new adjacency graph within the graph creation~(\ref{eq:graph}).

In other words, the $\rho_l$ function broadcasts the information of the nodes' neighborhoods (as determined by the adjacency on the previous step, $A^{(l)}$, and embedded in the local context), and, at each edge, creates a score of the possible adjacency as a linear combination of the nodes' features restricted to the existing structure.

\begin{table*}[tb]
  \sisetup{
    table-format = 1.4,
  }
  \caption{Comparison of GLN against deep generative models, GraphRNN~(G.RNN), Kronecker~(Kron.), and MMSB, on the Community ($C=2$ and $C=4$), on all sequences of Surf100 and Surf400, and Geometric Figures datasets.  The evaluation metric is MMD for degree~(D), cluster~(C), and orbits~(O) shown row-wise per method, where smaller numbers denote better performance.}
  \label{tab:mmd}
  \vspace{.1in}
  \scriptsize
  \setlength{\colsep}{5pt}
  \centering
  \begin{tabular}{@{ }l@{\hspace{5pt}}>{\itshape}l@{\hspace{5pt}}S@{\hspace{\colsep}}S@{\hspace{\colsep}}S@{\hspace{\colsep}}S@{\hspace{\colsep}}S@{\hspace{\colsep}}S@{\hspace{\colsep}}S@{\hspace{\colsep}}S@{\hspace{\colsep}}S@{\hspace{\colsep}}S@{\hspace{\colsep}}S@{\hspace{\colsep}}S@{\hspace{\colsep}}S@{\hspace{\colsep}}S@{\hspace{\colsep}}S@{\hspace{\colsep}}S@{\hspace{\colsep}}S@{ }}
    \toprule
    & & {\multirow{2}{*}{\textbf{C2}}} & {\multirow{2}{*}{\textbf{C4}}} & \multicolumn{7}{c}{\textbf{Surf400}} & \multicolumn{7}{c}{\textbf{Surf100}} & {\multirow{2}{*}{\textbf{Geo}}} \\
    \cmidrule{5-18}
    & & & & \textbf{T} & \textbf{EP} & \textbf{S} & \textbf{E} & \textbf{EH} & \textbf{O} & \textbf{A} & \textbf{T} & \textbf{EP} & \textbf{S} & \textbf{E} & \textbf{EH} & \textbf{O} & \textbf{A} & \\
    \midrule
    \multirow{3}{*}{\rotatebox{90}{GLN}} & D & 0.0121 & 0.0022 & 0.0001 & 0.0001 & 0.0001 & 0.0001 & 0.0001 & 0.0001 & 0.0016 & 0.0001 & 0.0001 & 0.0001 & 0.0001 & 0.0001 & 0.0001 & 0.0005 & 0.00621 \\
    & C & 0.0098 & 0.0026 & 0.0001 & 0.0001 & 0.0001 & 0.0001 & 0.0001 & 0.0001 & 0.0006 & 0.0001 & 0.0001 & 0.0001 & 0.0001 & 0.0001 & 0.0001 & 0.0003  & 0.000188\\
    & O & 0.6248 & 0.9952 & 0.0001 & 0.0001 & 0.0001 & 0.0001 & 0.0001 & 0.0001 & 0.0005 & 0.0001 & 0.0001 & 0.0001 & 0.0001 & 0.0001 & 0.0001 & 0.0002  & 0.00532\\[5pt]
    \multirow{3}{*}{\rotatebox{90}{G.RNN}} & D & 0.0027 & 0.2843 & 0.0287 & 0.0232 & 0.0303 & 0.0286 & 0.0436 & 0.0155 & 0.0388 & 0.0478 & 0.0506 & 0.1845 & 0.0664 & 0.0321 & 0.0880 & 0.0628 & 0.00228 \\
    & C & 0.0052 & 0.2272 & 1.6302 & 1.6690 & 1.7358 & 1.8362 & 1.8313 & 1.8057 & 1.7734 & 1.8271 & 1.0961 & 1.5689 & 1.7155 & 1.8379 & 1.9252 & 1.8962  & 0.0001 \\
    & O & 0.0033 & 1.9987 & 1.3684 & 1.3304 & 1.7337 & 1.5440 & 1.6709 & 1.5646 & 1.4736 & 0.4124 & 0.3705 & 0.8566 & 0.7786 & 0.9005 & 0.5702 & 1.5494  & 0.00150 \\[5pt]
    \multirow{3}{*}{\rotatebox{90}{Kron.}} & D & 1.0295 & 1.3741 & 0.9231 & 0.8922 & 0.9301 & 0.8873 & 0.8890 & 0.8987 & 0.9028 & 0.7361 & 0.8012 & 0.7279 & 0.7453 & 0.6382 & 0.8655 & 0.8515 & 0.58170 \\
    & C & 1.2837 & 1.3962 & 1.7836 & 1.7955 & 1.8163 & 1.8791 & 1.8814 & 1.8123 & 1.8945 & 1.9098 & 1.7722 & 1.7869 & 1.8981 & 1.9020 & 1.9297 & 1.9063  & 0.38149\\
    & O & 1.1846 & 1.3283 & 1.5621 & 1.5875 & 1.7834 & 1.6223 & 1.7027 & 1.6928 & 1.6338 & 0.4299 & 0.6013 & 0.5674 & 0.5655 & 0.6731 & 0.5827 & 1.3719  & 0.50516 \\[5pt]
    \multirow{3}{*}{\rotatebox{90}{MMSB}} & D & 1.7610 & 1.7457 & 1.1160 & 1.0256 & 1.1054 & 1.0513 & 1.0628 & 1.0589 & 1.0435 & 1.0124 & 1.0122 & 0.9940 & 1.0583 & 0.9334 & 1.1648 & 0.9825 & 0.61630 \\
    & C & 1.8817 & 1.9876 & 1.9987 & 1.9916 & 1.9985 & 1.9959 & 1.9975 & 1.9969 & 1.9951 & 1.9526 & 1.9417 & 1.9642 & 1.9744 & 1.9489 & 1.9332 & 1.9369  & 0.28545 \\
    & O & 1.4524 & 1.5095 & 1.7501 & 1.7851 & 1.8254 & 1.7863 & 1.7606 & 1.7480 & 1.7286 & 0.4303 & 0.7118 & 0.2466 & 0.6605 & 0.1209 & 0.7368 & 1.1789  & 0.60656 \\
    \bottomrule
  \end{tabular}
\end{table*}

\section{Learning Framework}
\label{sec:training}
We are assuming that we have a family of undirected graphs, $\mathcal{G} = \{ G_i \}_i$, that have a particular structure pattern that we are interested in.  
We will use each of the graphs, $G_i = (V_i, A_i)$, to learn the parameters, $\Theta$, of our model that minimize the loss function~(\ref{eq:loss}) on each of them.  
The structure of each graph is used as ground truth, $A_i^*=A_i$.  
The graph is predicted by the set of node embeddings, $\lambda_l$~(\ref{eq:local}), and the adjacency prediction, $\rho_l$~(\ref{eq:graph}), functions that depend on the weight matrices (\ie, $\Theta$) that are learnable, defined in Section~\ref{sec:gln}.

Our input comprises the vertices, $H^{(0)} = V_i$, and some structure for training.  
In our experiments, we used the identity, $A^{(0)} = I$.  
However, other structures can be used as well.  
In the following, we describe our learning framework to obtain the parameters $\theta_l \in \Theta$ of our functions for every $l$.  
For brevity, we will omit the parameters on the losses and in their functions.

Given the combinations of pairs of vertices on a graph, the total number of pairs with an edge (positive class) is, commonly, fewer than pairs without an edge (negative class).  
In order to handle the imbalance between the two binary classes (edge, no edge), we used the HED-loss function \citep{Xie2015} that is a class-balanced cross-entropy function.  
Then we consider the edge-class objective function as
\begin{equation}
  \mathcal{L}_{\mathit{c}} = -\beta \sum_{i \in Y_+} \log P \left( A_i^o \right) - (1-\beta) \sum_{j \in Y_-} \log P \left( A_j^o \right),
\end{equation}
where $A_i^{o}$ is the indexed predicted edge (output) for the $i$th pair of vertices. 
The proportion of positive (edge) and negative (no edge) pairs of vertices on the $A^*$ graph are $\beta = |Y_+|/|Y|$ and $1-\beta = |Y_-|/|Y|$, where $Y = Y_+ \cup Y_-$.
And $P(\cdot)$ is the probability of a pair of vertices to have an edge, predicted at the last layer~$L$, such that
\begin{equation}
  P \left( A_i^o \right) =  A_i^{(L)}.
\end{equation}
Individually penalizing the (class) prediction of each edge is not enough to model the structure of the graph.  
Hence, we compare the whole structure of the predicted graph, $A^{o}$, with its ground truth, $A^{*}$.  
By treating the edges on the adjacency matrices as regions on an image, we maximize the intersection-over-union \citep{Milletari2016} of the structural regions.  
Then we consider the objective function, 
\begin{equation}
  \mathcal{L}_{\mathit{s}} = 1 - \frac{2 |{A^{o} \cap A^*}|}{|A^{o}|^2 + |A^*|^2} = 1 - \frac{2 \sum\limits_{i,j} A^{o}_{i,j} A^{*}_{i,j}}{\sum\limits_{i,j} (A^{o}_{i,j})^2 + \sum\limits_{i,j}(A^{*}_{i,j})^2}.
\end{equation}
Finally, we aim to minimize the total loss that is the sum of all of the previous ones, defined by
\begin{equation}
\label{eq:loss}
\mathcal{L} = \psi_1 \mathcal{L}_{\mathit{c}} + \psi_2 \mathcal{L}_{\mathit{s}},
\end{equation}
where $\psi_1$ and $\psi_2$ are hyper-parameters that define the contribution of each loss to the learning process.

\section{Results and Discussion}
\label{sec:results}
In this work, we evaluate our model as an edge classifier, and simulate its performance as a graph generator by inputting noise as features and predicting on them.  
We perform experiments on three synthetic datasets that consist of images with Geometric Figures for segmentation, 3D surface function, and Community dataset (see Appendices~\ref{sec:figs}, \ref{sec:surfs}, and~\ref{sec:community}, respectively).
For our experiments, we used $80\%$ of the graphs in each dataset for training, and test on the rest.
Our evaluation metric is the Maximum Mean Discrepancy (MMD) measure \citep{You2018}, which measures the Wasserstein distance over three statistics of the graphs: degree (Deg), clustering coefficients (Clus), and orbits (Orb).  

We report our results contrasted against existing methods on Table~\ref{tab:mmd}. 
Additionally, we show more experiments using accuracy (Acc), intersection-over-union (IoU), and dice coefficient (Dice) in Appendix~\ref{sec:tabpred}.

\begin{figure}
  \centering
  \begin{tikzpicture}%
  \begin{axis}[%
    width=3.5cm,
    height=3.5cm,
    xlabel={Steps}, 
    ylabel=Dissimilarity,
    xtick={1,3,...,8},
    x tick label style={/pgf/number format/precision=0},
    enlarge y limits=false,
    ymax=2,
    legend pos=outer east,
    legend style={
      cells={anchor=west},
      font=\tiny,
    },
    cycle list/Dark2,
    cycle multiindex list={
      mark list\nextlist
      Dark2\nextlist
    },
  ]%
  \addplot table[x=deeplayer, y=mean_degree, col sep=comma]{img/deep_layer.csv};%
  \addlegendentry{Deg}%
  \addplot table[x=deeplayer, y=mean_clusteting, col sep=comma]{img/deep_layer.csv};%
  \addlegendentry{Clus}%
  \addplot table[x=deeplayer, y=mean_orbit, col sep=comma]{img/deep_layer.csv};%
  \addlegendentry{Orb}%
  \end{axis}%
  \end{tikzpicture}%
 \caption{%
  Results of the dissimilarity (MMD) between the prediction and ground truth (smaller values are better) while varying the number of recurrent steps, on the 3D Surface dataset (Surf400).
}
\label{fig:deep_layer}
\end{figure}
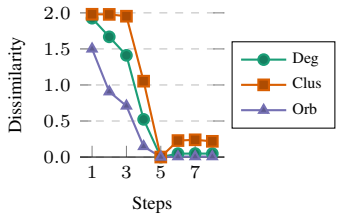

Knowing the depth of the recursive model (\ie, the number of iterations)  is not a trivial task since we must find a trade-off between the efficiency and effectiveness of the model.
In Fig.~\ref{fig:deep_layer}, we show the dissimilarity metrics (MMD) while varying the number of applications of our proposed block on the 3D Surface dataset.
According to our experiment, using five recurrent steps provides the right trade-off.

\begin{table}[tb]
  \centering
  \sisetup{
    table-format = 1.5,
  }
  \newrobustcmd{\B}{\bfseries}
  \caption{Ablation of GLN using Geometric Figures. Note, in the first three metrics, high values are better, and opposite in the rest.}
  \label{tab:ablation}
  \vspace{.15in}
  \scriptsize
  \setlength{\colsep}{5pt}
  \begin{tabular}{%
      @{ }S@{\hspace{\colsep}}%
      S@{\hspace{\colsep}}%
      S@{\hspace{\colsep}}%
      S@{\hspace{2pt}}%
      S@{\hspace{\colsep}}
      S@{\hspace{\colsep}}
      S@{\hspace{\colsep}}
      S@{\hspace{\colsep}}
      S@{\hspace{\colsep}}
      S@{ }
    }
    \toprule
    \multicolumn{3}{c}{\textbf{Losses}} & & \multicolumn{5}{c}{\textbf{Metrics}} \\
    \cmidrule{1-3} \cmidrule{5-10}
    \textbf{IoU} & \textbf{HED} & \textbf{Reg} & & \textbf{Acc}$\uparrow$ & \textbf{IoU}$\uparrow$  & \textbf{Dice}$\uparrow$ & \textbf{Deg}$\downarrow$ & \textbf{Clus}$\downarrow$ & \textbf{Orb}$\downarrow$\\
    \midrule
    {--} & {\checkmark} & {--} & & 0.99969346 & 0.97466326 & 0.9871683 & 0.006767745310741997 & 0.00111111291736 & 0.10686012485836936\\
    {--} & {\checkmark} & {\checkmark} & & 0.99969586 & 0.97489115 & 0.9872487 & 0.0064840807136821434 & 0.00101826180551 & 0.09724425232672966\\
    {\checkmark} & {--} & {--} & & 0.799676 & 0.052401826 & 0.09958515 & 1.8624341298943576 & 1.998031214905416 & 0.9827227274368951\\
    {\checkmark} & {--} & {\checkmark} & & 0.89377505 & 0.09526738 & 0.17396164 & 1.7689470157408824 & 1.9491198237082703 & 1.1861635441598561\\
    {\checkmark} & {\checkmark} & {--} & & 0.9996951 & \B 0.97489555 & \B 0.987233 & 0.006273512387788255 & 0.00019173764942 & 0.061865321693253476\\
    {\checkmark} & {\checkmark} & {\checkmark} & & \B 0.99969786 & 0.97488553 & 0.98724604 & \B 0.006215694036369642 & \B 0.000188182646141 & \B 0.005322178684469159\\
    \bottomrule
  \end{tabular}
\end{table}

\begin{figure}
\centering
\centering
\begin{tikzpicture}%
\pgfplotsset{every non boxed x axis/.style={}}
\begin{groupplot}[
  width=.45\linewidth,
  height=2.5cm,
  minimal plot grid,
  y axis line style={opacity=1},
  group style={
    group name=plot,
    group size=2 by 2,
    xticklabels at=edge bottom,
    xlabels at=edge bottom,
    vertical sep=2.5pt,
    horizontal sep=1.5cm,
    x descriptions at=edge bottom,
  },
  xlabel=Proportion of edges, 
  ylabel=Distance,
  xtick={0.2,.4,...,1.0},
  legend style={
    font=\tiny,
    cells={anchor=west},
    legend columns=3,
  },
  cycle list/Dark2,
  y tick label style={
    /pgf/number format/precision=2,
  },
  enlarge y limits=false,
  enlarge x limits=false,
]%
\nextgroupplot[
  ymin=.3,
  ymax=1.2,
  ytick={.6, .8,..., 1.2},
  yticklabels={0.6, 0.8, ..., 1.2},
  axis x line=top, 
  axis y discontinuity=parallel,
  axis y line=right,
  ylabel=,
  legend to name=legend,
]
\addplot table[x=edgeproportion, y=mean_clusteting, col sep=comma]{img/robustness_C4.csv};%
\addplot table[x=edgeproportion, y=mean_degree, col sep=comma]{img/robustness_C4.csv};%
\addplot table[x=edgeproportion, y=mean_orbit, col sep=comma]{img/robustness_C4.csv};%
\legend{Clus,Deg,Orb}%
\nextgroupplot[
  ymin=.3,
  ymax=1.2,
  ytick={.6, .8, ..., 1.2},
  yticklabels={0.6, 0.8, ..., 1.2},
  group/y descriptions at=edge right,
  axis x line=top, 
  axis y discontinuity=parallel,
  axis y line=right,
  ylabel=,
]
\addplot table[x=edgeproportion, y=mean_clusteting, col sep=comma]{img/robustness_C2.csv};%
\addplot table[x=edgeproportion, y=mean_degree, col sep=comma]{img/robustness_C2.csv};%
\addplot table[x=edgeproportion, y=mean_orbit, col sep=comma]{img/robustness_C2.csv};%
\nextgroupplot[
  ymin=0.002,
  ymax=0.0041,
  ytick={0.0023, 0.003, 0.0037},
  axis x line=bottom,
  axis y line=left,
]
\addplot table[x=edgeproportion, y=mean_clusteting, col sep=comma]{img/robustness_C4.csv};%
\addplot table[x=edgeproportion, y=mean_degree, col sep=comma]{img/robustness_C4.csv};%
\nextgroupplot[
  ymin=0.007,
  ymax=0.020,
  ytick={0.009,0.013,0.017},
  axis x line=bottom,
  axis y line=left,
]
\addplot table[x=edgeproportion, y=mean_clusteting, col sep=comma]{img/robustness_C2.csv};%
\addplot table[x=edgeproportion, y=mean_degree, col sep=comma]{img/robustness_C2.csv};%
\end{groupplot}%
\path (plot c1r1.north) -- node[above, yshift=0cm]{\pgfplotslegendfromname{legend}} (plot c2r1.north);
\end{tikzpicture}%
\caption{MMD metrics on GLN when varying the input structure on Community $C=4$ (left) and $C=2$ (right).  The input corresponds to an adjacency matrix with different proportions of connections.}
\label{fig:robustness}
\vspace*{-15pt}
\end{figure}
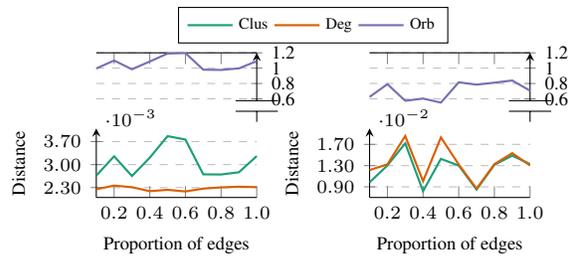

Additionally, in Table~\ref{tab:ablation}, we present an ablation analysis of our model's loss functions and regularization components on the Geometric Figures dataset. 
We emphasize a stable training and a fast convergence when we minimize both loss functions simultaneously.

Finally, we examined the robustness to structural inputs by randomly changing the proportion of the initial connections (\ie, $10\%$, $20\%$, $\dots$, $100\%$) in our input $A^{(0)}$.  
Fig.~\ref{fig:robustness} shows the average results (of five executions) of this experiment on the Community ($C=2$, and $C=4$).  
We obtained minimum variation on the prediction capabilities of the network.  
Hence, the best option is to select a minimal graph as input, \ie, the identity matrix. 
We present our models' qualitative results on the different databases in Appendices~\ref{sec:pred_surf}, \ref{sec:pred_comm}, and~\ref{sec:pred_geo}.

\section{Conclusions}
\label{sec:conclusion}
We proposed a simple yet effective method to predict the structure of a set of vertices.  
Our method works by learning node embedding and adjacency prediction functions and chaining them.  
This process produces expected embeddings which are used to obtain the most probable adjacency.  
We encode this process into the neural network architecture.  
Our experiments demonstrate the prediction capabilities of our model on three databases with structures with different features (the communities are densely connected on some parts, and sparse on others, while the images are connected with at most four neighbors).  
Further experiments are necessary to evaluate the robustness of the proposed method on larger graphs, with more features and more challenging structures.

\section*{Acknowledgements}

This work was financed in part by the S\~{a}o Paulo Research Foundation (FAPESP) under grants No.~2016/19947-6 and No.~2017/16597-7,  the Brazilian National Council for Scientific and Technological Development (CNPq) under grant No.~307425/2017-7, and the Coordenação de
Aperfeiçoamento de Pessoal de Nível Superior---Brasil (CAPES)---Finance Code~001.  We acknowledge the support of NVIDIA Corporation for the donation of a Titan~X Pascal GPU used in this research.


\bibliographystyle{icml2019}
\bibliography{abrv,icml2019_workshop}

\begin{thebibliography}{31}
\providecommand{\natexlab}[1]{#1}
\providecommand{\url}[1]{\texttt{#1}}
\expandafter\ifx\csname urlstyle\endcsname\relax
  \providecommand{\doi}[1]{doi: #1}\else
  \providecommand{\doi}{doi: \begingroup \urlstyle{rm}\Url}\fi

\bibitem[dgl()]{dgl.ai}
Deep graph library.
\newblock URL \url{https://www.dgl.ai/}.

\bibitem[Bai et~al.(2019)Bai, Ding, Bian, Chen, Sun, and Wang]{Bai2019}
Bai, Y., Ding, H., Bian, S., Chen, T., Sun, Y., and Wang, W.
\newblock {SimGNN}: A neural network approach to fast graph similarity
  computation.
\newblock In \emph{{ACM} Inter. Conf. Web Search Data Min. ({WSDM})}, WSDM '19,
  pp.\  384--392, New York, NY, USA, 2019. ACM.
\newblock ISBN 978-1-4503-5940-5.

\bibitem[Battaglia et~al.(2018)Battaglia, Hamrick, Bapst, Sanchez-Gonzalez,
  Zambaldi, Malinowski, Tacchetti, Raposo, Santoro, Faulkner,
  et~al.]{Battaglia2018}
Battaglia, P.~W., Hamrick, J.~B., Bapst, V., Sanchez-Gonzalez, A., Zambaldi,
  V., Malinowski, M., Tacchetti, A., Raposo, D., Santoro, A., Faulkner, R.,
  et~al.
\newblock Relational inductive biases, deep learning, and graph networks.
\newblock \emph{arXiv}, \penalty0 (1806.01261v3), 2018.

\bibitem[Berg et~al.(2018)Berg, Kipf, and Welling]{Berg2018}
Berg, R. v.~d., Kipf, T.~N., and Welling, M.
\newblock Graph convolutional matrix completion.
\newblock \emph{{ACM} Conf. Knowl. Discov. Data Min. ({ACM SIGKDD})}, 2018.

\bibitem[Bojchevski et~al.(2018)Bojchevski, Shchur, Z{\"u}gner, and
  G{\"u}nnemann]{Bojchevski2018}
Bojchevski, A., Shchur, O., Z{\"u}gner, D., and G{\"u}nnemann, S.
\newblock {NetGAN}: Generating graphs via random walks.
\newblock In \emph{Inter. Conf. Mach. Learn. ({ICML})}, 2018.

\bibitem[Bresson \& Laurent(2018)Bresson and Laurent]{Bresson2018}
Bresson, X. and Laurent, T.
\newblock Residual gated graph convnets, 2018.
\newblock URL \url{https://openreview.net/forum?id=HyXBcYg0b}.

\bibitem[De~Cao \& Kipf(2018)De~Cao and Kipf]{DeCao2018}
De~Cao, N. and Kipf, T.
\newblock {MolGAN}: An implicit generative model for small molecular graphs.
\newblock \emph{arXiv}, \penalty0 (1805.11973), 2018.

\bibitem[Defferrard et~al.(2016)Defferrard, Bresson, and
  Vandergheynst]{Defferrard2016}
Defferrard, M., Bresson, X., and Vandergheynst, P.
\newblock Convolutional neural networks on graphs with fast localized spectral
  filtering.
\newblock In \emph{Adv. Neural Inf. Process. Sys. ({NeurIPS})}, pp.\
  3844--3852, USA, 2016. Curran Associates Inc.
\newblock ISBN 978-1-5108-3881-9.

\bibitem[Donnat et~al.(2018)Donnat, Zitnik, Hallac, and Leskovec]{Donnat2018}
Donnat, C., Zitnik, M., Hallac, D., and Leskovec, J.
\newblock Learning structural node embeddings via diffusion wavelets.
\newblock In \emph{{ACM} Conf. Knowl. Discov. Data Min. ({ACM SIGKDD})}, pp.\
  1320--1329. ACM, 2018.

\bibitem[Gilmer et~al.(2017)Gilmer, Schoenholz, Riley, Vinyals, and
  Dahl]{Gilmer2017}
Gilmer, J., Schoenholz, S.~S., Riley, P.~F., Vinyals, O., and Dahl, G.~E.
\newblock Neural message passing for quantum chemistry.
\newblock In Precup, D. and Teh, Y.~W. (eds.), \emph{Inter. Conf. Mach. Learn.
  ({ICML})}, volume~70 of \emph{Proceedings of Machine Learning Research}, pp.\
   1263--1272, International Convention Centre, Sydney, Australia, 06--11 Aug
  2017. PMLR.

\bibitem[Gori et~al.(2005)Gori, Monfardini, and Scarselli]{Gori2005}
Gori, M., Monfardini, G., and Scarselli, F.
\newblock A new model for learning in graph domains.
\newblock In \emph{{IEEE} Inter. Joint Conf. Neural Netw. ({IJCNN})}, volume~2,
  pp.\  729--734. IEEE, 2005.

\bibitem[Grover et~al.(2018)Grover, Zweig, and Ermon]{Grover2018}
Grover, A., Zweig, A., and Ermon, S.
\newblock {Graphite}: Iterative generative modeling of graphs.
\newblock \emph{arXiv}, \penalty0 (1803.10459v3), 2018.

\bibitem[Kearnes et~al.(2019)Kearnes, Li, and Riley]{Kearnes2019}
Kearnes, S., Li, L., and Riley, P.
\newblock Decoding molecular graph embeddings with reinforcement learning.
\newblock \emph{arXiv}, \penalty0 (1904.08915), 2019.

\bibitem[Kingma \& Welling(2014)Kingma and Welling]{Kingma2014}
Kingma, D.~P. and Welling, M.
\newblock Auto-encoding variational bayes.
\newblock \emph{Inter. Conf. Learn. Represent. ({ICLR})}, 1050:\penalty0 1,
  2014.

\bibitem[Kipf et~al.(2018)Kipf, Fetaya, Wang, Welling, and Zemel]{Kipf2018}
Kipf, T., Fetaya, E., Wang, K.-C., Welling, M., and Zemel, R.
\newblock Neural relational inference for interacting systems.
\newblock \emph{arXiv}, \penalty0 (1802.04687v2), 2018.

\bibitem[Kipf \& Welling(2017)Kipf and Welling]{Kipf2017}
Kipf, T.~N. and Welling, M.
\newblock Semi-supervised classification with graph convolutional networks.
\newblock In \emph{Inter. Conf. Learn. Represent. ({ICLR})}, 2017.

\bibitem[Kusner et~al.(2017)Kusner, Paige, and
  Hern{\'a}ndez-Lobato]{Kusner2017}
Kusner, M.~J., Paige, B., and Hern{\'a}ndez-Lobato, J.~M.
\newblock Grammar variational autoencoder.
\newblock In \emph{Inter. Conf. Mach. Learn. ({ICML})}, pp.\  1945--1954, 2017.

\bibitem[Li et~al.(2016)Li, Zemel, and Brockschmidt]{Li2016}
Li, Y., Zemel, R., and Brockschmidt, M.~a.
\newblock Gated graph sequence neural networks.
\newblock In \emph{Inter. Conf. Learn. Represent. ({ICLR})}, April 2016.

\bibitem[Li et~al.(2018)Li, Vinyals, Dyer, Pascanu, and Battaglia]{Li2018}
Li, Y., Vinyals, O., Dyer, C., Pascanu, R., and Battaglia, P.
\newblock Learning deep generative models of graphs.
\newblock \emph{Inter. Conf. Learn. Represent. ({ICLR})}, 2018.

\bibitem[Marcheggiani \& Titov(2017)Marcheggiani and Titov]{Marcheggiani2017}
Marcheggiani, D. and Titov, I.
\newblock Encoding sentences with graph convolutional networks for semantic
  role labeling.
\newblock In \emph{Conf. Empir. Methods Nat. Lang. Process. ({EMNLP})}, pp.\
  1506--1515, Copenhagen, Denmark, September 2017. Association for
  Computational Linguistics.

\bibitem[Milletari et~al.(2016)Milletari, Navab, and Ahmadi]{Milletari2016}
Milletari, F., Navab, N., and Ahmadi, S.-A.
\newblock {V-net}: Fully convolutional neural networks for volumetric medical
  image segmentation.
\newblock In \emph{{IEEE} Inter. Conf. 3D Vis. ({3DV})}, pp.\  565--571. IEEE,
  2016.

\bibitem[Monti et~al.(2017)Monti, Bronstein, and Bresson]{Monti2017}
Monti, F., Bronstein, M., and Bresson, X.
\newblock Geometric matrix completion with recurrent multi-graph neural
  networks.
\newblock In Guyon, I., Luxburg, U.~V., Bengio, S., Wallach, H., Fergus, R.,
  Vishwanathan, S., and Garnett, R. (eds.), \emph{Adv. Neural Inf. Process.
  Sys. ({NeurIPS})}, pp.\  3697--3707. Curran Associates, Inc., 2017.

\bibitem[Scarselli et~al.(2009)Scarselli, Gori, Tsoi, Hagenbuchner, and
  Monfardini]{Scarselli2009}
Scarselli, F., Gori, M., Tsoi, A.~C., Hagenbuchner, M., and Monfardini, G.
\newblock Computational capabilities of graph neural networks.
\newblock \emph{{IEEE} Trans. Neural Netw.}, 20\penalty0 (1):\penalty0 81--102,
  2009.

\bibitem[Schlichtkrull et~al.(2018)Schlichtkrull, Kipf, Bloem, van den Berg,
  Titov, and Welling]{Schlichtkrull2018}
Schlichtkrull, M., Kipf, T.~N., Bloem, P., van den Berg, R., Titov, I., and
  Welling, M.
\newblock Modeling relational data with graph convolutional networks.
\newblock In Gangemi, A., Navigli, R., Vidal, M.-E., Hitzler, P., Troncy, R.,
  Hollink, L., Tordai, A., and Alam, M. (eds.), \emph{Semantic Web Conf.
  ({ESWC})}, pp.\  593--607, Cham, 2018. Springer International Publishing.

\bibitem[Simonovsky \& Komodakis(2018)Simonovsky and Komodakis]{Simonovsky2018}
Simonovsky, M. and Komodakis, N.
\newblock {GraphVAE}: Towards generation of small graphs using variational
  autoencoders.
\newblock In \emph{Int. Conf. Artif. Neural Netw. ({ICANN})}, pp.\  412--422.
  Springer, 2018.

\bibitem[Sohn et~al.(2015)Sohn, Lee, and Yan]{Sohn2015}
Sohn, K., Lee, H., and Yan, X.
\newblock Learning structured output representation using deep conditional
  generative models.
\newblock In Cortes, C., Lawrence, N.~D., Lee, D.~D., Sugiyama, M., and
  Garnett, R. (eds.), \emph{Adv. Neural Inf. Process. Sys. ({NeurIPS})}, pp.\
  3483--3491. Curran Associates, Inc., 2015.

\bibitem[Watts(1999)]{Watts1999}
Watts, D.~J.
\newblock Networks, dynamics, and the small-world phenomenon.
\newblock \emph{Amer. J. Soc.}, 105\penalty0 (2):\penalty0 493--527, 1999.

\bibitem[Xie \& Tu(2015)Xie and Tu]{Xie2015}
Xie, S. and Tu, Z.
\newblock Holistically-nested edge detection.
\newblock In \emph{{IEEE} Inter. Conf. Comput. Vis. ({ICCV})}, pp.\
  1395--1403, 2015.

\bibitem[Ying et~al.(2018)Ying, You, Morris, Ren, Hamilton, and
  Leskovec]{Ying2018}
Ying, Z., You, J., Morris, C., Ren, X., Hamilton, W., and Leskovec, J.
\newblock Hierarchical graph representation learning with differentiable
  pooling.
\newblock In \emph{Adv. Neural Inf. Process. Sys. ({NeurIPS})}, pp.\
  4800--4810, 2018.

\bibitem[You et~al.(2018)You, Ying, Ren, Hamilton, and Leskovec]{You2018}
You, J., Ying, R., Ren, X., Hamilton, W., and Leskovec, J.
\newblock {GraphRNN}: Generating realistic graphs with deep auto-regressive
  models.
\newblock In \emph{Inter. Conf. Mach. Learn. ({ICML})}, pp.\  5694--5703, 2018.

\bibitem[Zhang \& Chen(2018)Zhang and Chen]{Zhang2018a}
Zhang, M. and Chen, Y.
\newblock Link prediction based on graph neural networks.
\newblock In \emph{Adv. Neural Inf. Process. Sys. ({NeurIPS})}, 2018.

\end{thebibliography}

\appendix
\counterwithin{figure}{section}
\counterwithin{table}{section}

\twocolumn[%
\icmltitle{Graph Learning Network: A Structure Learning Algorithm\\
      {\normalsize SUPPLEMENTARY MATERIAL}}



\icmlsetsymbol{equal}{*}
\vskip 0.3in
]
\section{Datasets}
\subsection{Geometric Figures Dataset}
\label{sec:figs}
We made the Geometric Figures dataset for the task of image segmentation within a controlled environment. 
Segmentation is given by the connected components of the graph ground-truth.  
Here, we provide RGB images and their expected segmentations.

The Geometric Figures dataset contains $3000$ images of size $n \times n$, that are generated procedurally.\footnote{Code available at \url{https://gitlab.com/mipl/graph-learning-network}.}
Each image contains circles, rectangles, and lines (dividing the image into two parts).  
We also add white noise to the color intensity of the images to perturb and mixed their regions.

The geometrical figures are of different dimensions, within $[1, n]$, and positioned randomly on the image (taking care in maintaining the geometric figure).
There is no specific color for each geometric shape and their background.

For our experiments we use a version of dimension $n = 20$. 

\subsection{3D Surfaces Dataset}
\label{sec:surfs}
To evaluate our method we needed a highly structured dataset with intricate relations and with easily understandable features.  
Hence, we convert parts of 3D surfaces into a mesh by sampling them.  
Each point in the mesh is translated into a node of the graph, with its position as a feature vector.  
We have a generator\footnote{Code available at \url{https://gitlab.com/mipl/graph-learning-network}.} that creates different configurations for this dataset based on a number of nodes per surface, and transformation on it.  

We considered the following surfaces:
\begin{itemize}
  \item \textbf{Ellipsoid:} defined by the 3D-function $\frac{x^2}{a^2} + \frac{y^2}{b^2} + \frac{z^2}{c^2}= 1$, where the semi-axes are of lengths $a$, $b$, and $c$.
  \item \textbf{Elliptic hyperboloid:} defined by the 3D-function $\frac{x^2}{a^2} + \frac{y^2}{b^2} - \frac{z^2}{c^2}= 1$, where the semi-axes are of lengths $a$, $b$, and $c$.
  \item \textbf{Elliptic paraboloid:} defined by the 3D-function $\frac{x^2}{a^2} + \frac{y^2}{b^2} = z$, where $a$ and $b$ are the level of curvature in the $xz$ and $yz$ planes respectively.
  \item \textbf{Saddle:} defined by the 3D-function $\frac{x^2}{a^2} - \frac{y^2}{b^2} = z$, where $a$ and $b$ are the level of curvature in the $xz$ and $yz$ planes respectively.
  \item \textbf{Torus:} defined by the 3D-function $\left(\sqrt{x^2 + y^2} - R \right)^2 + z^2 = r^2$, where $R$ is the major radius and $r$ is the minor radius.
  \item \textbf{Another:} defined by the 3D-function $ h  \sin(\sqrt{x^2 + y^2}) = z$, where $h$ is the height above z-axis.
\end{itemize}
We generated $200$ versions of each surface by randomly applying a set of transformations (from scaling, translation, rotation, reflection, or shearing) to the curve, moreover, two versions of the Surface dataset were created, \textit{Surf100} and \textit{Surf400} that use $100$ and $400$ vertices per surface, respectively.

\subsection{Community Dataset}
\label{sec:community}
We perform experiments on a synthetic dataset (Community dataset) that comprises two sets with $C=2$ and $C=4$ communities with $40$ and $80$ vertices each, respectively, created with the caveman algorithm \citep{Watts1999}, where each community has $20$ people.  
Besides, Community $C=4$ and $C=2$ have 500 and 300 samples respectively.

\section{Architecture}
\label{sec:arch}
For our experiments, we used $80\%$ of the graphs in each dataset for training, and test on the rest.
For both models, we use the following settings.  Our activation functions, $\sigma_l$, are sigmoid for all layers, except for the Eq.~\ref{eq:global} where $\sigma_l$ is a hyperbolic tangent.  
We use $L=5$ layers to extract the final adjacency and embeddings.  
The feature dimension, $d_l$, is $32$ for all layers.  
The learning rate is set \num{e-5} for the Community dataset, and in the rest of datasets, the learning rate is set \num{5e-6}.
Additionally, the number of epochs changes depending on the experiment. Thus in the experiments of Communities, Surfaces and Geometrical Figures we use $150$, $200$ and $150$ times respectively and, the number of kernel using is $k=3$.
To convert the prediction of the adjacency into a binary edge, we use a fixed threshold of $\epsilon = 0.5$.  
The hyper-parameters in our loss function~(\ref{eq:loss}) are $\psi_1=1$ and $\psi_2=1$.  In our experiments, we did not needed the regularization our GLN model.  
Finally, for training, we used the ADAM optimization algorithm on Nvidia GTX Titan X GPU with $12$\,GB of memory.

\section{More Measure of Prediction}
\label{sec:tabpred}
Unlike Table~\ref{tab:mmd}, where dissimilarity measures are used, such as our metric evaluation on graphs, in Table~\ref{tab:pred} we present similarity measures such as accuracy~(Acc), intersection-over-union~(IoU), Recall~(Rec), and Precision~(Prec).

\begin{table*}[tb]
  \centering
  \sisetup{
    table-format = 1.3,
  }
  \caption{%
    Comparison of GLN, on the Community ($C=2$ and $C=4$), on all sequences of Surf100 and Surf400, and Geometric Figures datasets.  
    The evaluation metric are accuracy~(Acc), intersection-over-union~(IoU), Recall~(Rec), and Precision~(Prec) shown row-wise per method, where larger numbers denote better performance.
  }
  \label{tab:pred}
  \vspace{.15in}
  \scriptsize
  \setlength{\colsep}{5pt}
  \begin{tabular}{@{ }l@{\hspace{5pt}}>{\itshape}l@{\hspace{5pt}}S@{\hspace{\colsep}}S@{\hspace{\colsep}}S@{\hspace{\colsep}}S@{\hspace{\colsep}}S@{\hspace{\colsep}}S@{\hspace{\colsep}}S@{\hspace{\colsep}}S@{\hspace{\colsep}}S@{\hspace{\colsep}}S@{\hspace{\colsep}}S@{\hspace{\colsep}}S@{\hspace{\colsep}}S@{\hspace{\colsep}}S@{\hspace{\colsep}}S@{\hspace{\colsep}}S@{\hspace{\colsep}}S@{ }}
    \toprule
    & & {\multirow{2}{*}{\textbf{C2}}} & {\multirow{2}{*}{\textbf{C4}}} & \multicolumn{7}{c}{\textbf{Surf400}} & \multicolumn{7}{c}{\textbf{Surf100}} & {\multirow{2}{*}{\textbf{Geo}}} \\
    \cmidrule{5-18}
    & & & & \textbf{T} & \textbf{EP} & \textbf{S} & \textbf{E} & \textbf{EH} & \textbf{O} & \textbf{A} & \textbf{T} & \textbf{EP} & \textbf{S} & \textbf{E} & \textbf{EH} & \textbf{O} & \textbf{A} \\
    \midrule
    \multirow{3}{*}{\rotatebox{90}{GLN}} & Acc & 0.997 & 0.997 & 0.999 & 0.999 & 0.999 & 0.999 & 0.999 & 0.999 & 0.997 & 0.999 & 0.999 & 0.999 & 0.999 & 0.999 & 0.999 & 0.993 & 0.999  \\
    & IoU & 0.993 & 0.992 & 0.991 & 0.982 & 0.999 & 0.981 & 0.989 & 0.999 & 0.865 & 0.999 & 0.999 & 0.999 & 0.999 & 0.999 & 0.999 & 0.877 & 0.974 \\
    & Rec & 0.994 & 0.997 & 0.999 & 0.999 & 0.999 & 0.999 & 0.999 & 0.999 & 0.928 & 0.999 & 0.999 & 0.999 & 0.999 & 0.999 & 0.999 & 0.934 & 0.986 \\
    & Prec & 0.997 & 0.997 & 0.991 & 0.982 & 0.999 & 0.981 & 0.989 & 0.999 & 0.927 & 0.999 & 0.999 & 0.999 & 0.999 & 0.999 & 0.999 & 0.934 & 0.976\\
    \bottomrule
  \end{tabular}
\end{table*}

\section{Prediction of 3D Surface}
\label{sec:pred_surf}
In Fig.~\ref{fig:pred_surface}, we show the qualitative result of GLN for the 3D Surface dataset. 
We show the prediction on the elliptic hyperboloid, elliptic paraboloid, torus, saddle, and ellipsoid, all using $100$ nodes (Surf100).
We normalized the graphs (\wrt scale and translation) for better visualization.
Besides, the red edges represent false negatives (\ie, not predicted edges) and black edges are correctly predicted ones.

\setlength{\subfigwidth}{.2\textwidth}
\setlength{\subfigheight}{.75\textwidth}
\begin{figure*}
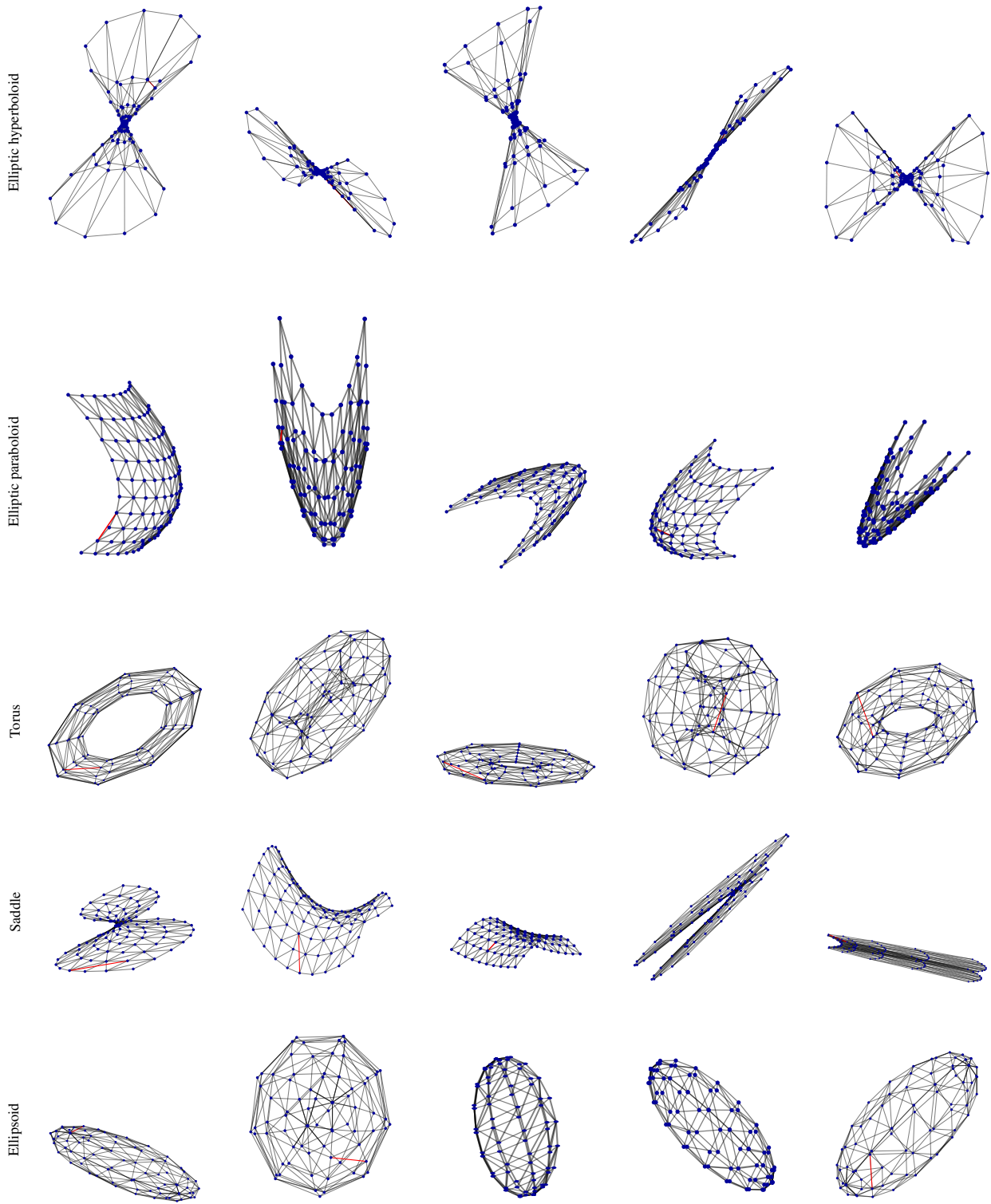
%
  \centering%
  \setlength{\subfigwidth}{0.19\textwidth}
  \begin{tabular}{@{}c@{ }@{}c@{}@{}c@{}@{}c@{}@{}c@{}@{}c@{}}%
    \raisebox{.35\subfigwidth}{\rotatebox{90}{\scriptsize Elliptic hyperboloid}} & 
    \resizebox{\subfigwidth}{!}{\graphSurface{img/elliptic_hyperboloid_1_edge_norm.csv}{img/elliptic_hyperboloid_1_node_norm.csv}{0.25}} &
    \resizebox{\subfigwidth}{!}{\graphSurface{img/elliptic_hyperboloid_3_edge_norm.csv}{img/elliptic_hyperboloid_3_node_norm.csv}{0.25}} &
    \resizebox{\subfigwidth}{!}{\graphSurface{img/elliptic_hyperboloid_7_edge_norm.csv}{img/elliptic_hyperboloid_7_node_norm.csv}{0.25}} &
    \resizebox{\subfigwidth}{!}{\graphSurface{img/elliptic_hyperboloid_21_edge_norm.csv}{img/elliptic_hyperboloid_21_node_norm.csv}{0.35}} &
    \resizebox{\subfigwidth}{!}{\graphSurface{img/elliptic_hyperboloid_8_edge_norm.csv}{img/elliptic_hyperboloid_8_node_norm.csv}{0.35}} \\ 
    \raisebox{.35\subfigwidth}{\rotatebox{90}{\scriptsize Elliptic paraboloid}} &
    \resizebox{\subfigwidth}{!}{\graphSurface{img/elliptic_paraboloid_0_edge_norm.csv}{img/elliptic_paraboloid_0_node_norm.csv}{0.15}} &
    \resizebox{\subfigwidth}{!}{\graphSurface{img/elliptic_paraboloid_1_edge_norm.csv}{img/elliptic_paraboloid_1_node_norm.csv}{0.15}} &
    \resizebox{\subfigwidth}{!}{\graphSurface{img/elliptic_paraboloid_4_edge_norm.csv}{img/elliptic_paraboloid_4_node_norm.csv}{0.2}} &
    \resizebox{\subfigwidth}{!}{\graphSurface{img/elliptic_paraboloid_5_edge_norm.csv}{img/elliptic_paraboloid_5_node_norm.csv}{0.15}} &
    \resizebox{\subfigwidth}{!}{\graphSurface{img/elliptic_paraboloid_9_edge_norm.csv}{img/elliptic_paraboloid_9_node_norm.csv}{0.15}} \\ 
    \raisebox{.35\subfigwidth}{\rotatebox{90}{\scriptsize Torus}} &
    \resizebox{\subfigwidth}{!}{\graphSurface{img/torus_0_edge_norm.csv}{img/torus_0_node_norm.csv}{0.15}} &
    \resizebox{\subfigwidth}{!}{\graphSurface{img/torus_11_edge_norm.csv}{img/torus_11_node_norm.csv}{0.15}} &
    \resizebox{\subfigwidth}{!}{\graphSurface{img/torus_9_edge_norm.csv}{img/torus_9_node_norm.csv}{0.2}} &
    \resizebox{\subfigwidth}{!}{\graphSurface{img/torus_2_edge_norm.csv}{img/torus_2_node_norm.csv}{0.15}} &
    \resizebox{\subfigwidth}{!}{\graphSurface{img/torus_51_edge_norm.csv}{img/torus_51_node_norm.csv}{0.15}} \\ 
    \raisebox{.35\subfigwidth}{\rotatebox{90}{\scriptsize Saddle}} &
    \resizebox{\subfigwidth}{!}{\graphSurface{img/saddle_3_edge_norm.csv}{img/saddle_3_node_norm.csv}{0.16}} &
    \resizebox{\subfigwidth}{!}{\graphSurface{img/saddle_6_edge_norm.csv}{img/saddle_6_node_norm.csv}{0.17}} &
    \resizebox{\subfigwidth}{!}{\graphSurface{img/saddle_9_edge_norm.csv}{img/saddle_9_node_norm.csv}{0.15}} &
    \resizebox{\subfigwidth}{!}{\graphSurface{img/saddle_28_edge_norm.csv}{img/saddle_28_node_norm.csv}{0.2}} &
    \resizebox{\subfigwidth}{!}{\graphSurface{img/saddle_22_edge_norm.csv}{img/saddle_22_node_norm.csv}{0.15}} \\ 
    \raisebox{.35\subfigwidth}{\rotatebox{90}{\scriptsize Ellipsoid}} &
    \resizebox{\subfigwidth}{!}{\graphSurface{img/ellipsoid_1_edge_norm.csv}{img/ellipsoid_1_node_norm.csv}{0.16}} &
    \resizebox{\subfigwidth}{!}{\graphSurface{img/ellipsoid_61_edge_norm.csv}{img/ellipsoid_61_node_norm.csv}{0.17}} &
    \resizebox{\subfigwidth}{\subfigwidth}{\graphSurface{img/ellipsoid_5_edge_norm.csv}{img/ellipsoid_5_node_norm.csv}{0.15}} &
    \resizebox{\subfigwidth}{!}{\graphSurface{img/ellipsoid_6_edge_norm.csv}{img/ellipsoid_6_node_norm.csv}{0.2}} &
    \resizebox{\subfigwidth}{!}{\graphSurface{img/ellipsoid_8_edge_norm.csv}{img/ellipsoid_8_node_norm.csv}{0.15}} \\ 
  \end{tabular}%
  \caption{Results on 3D Surface dataset predictions for the proposed methods, and the learned latent space, used for build adjacency matrix in the prediction.  The blue edges represent false negatives (\ie, not predicted edges), red edges represent false positives (\ie, additional predicted edges), and black edges are correctly predicted ones.  The graphs were normalized (\wrt scale and translation) for better visualization.}
  \label{fig:pred_surface}%
\end{figure*}

\section{Prediction of Community}
\label{sec:pred_comm}
In Fig.~\ref{fig:pred_community}, we predict the adjacency matrix the of Community dataset on two and four communities, $C=2$ and $C=4$ respectively (even rows).
Note, our node embedding obtained after apply the  $\lambda_l$ function, shows a good grouping of individuals in the hyperspace (odd rows).
Furthermore, the red edges represent false negatives (\ie, not predicted edges), and black edges are correctly predicted ones.

\begin{figure*}%
  \centering%
  \setlength{\subfigwidth}{.12\linewidth}
  \begin{tabular}{@{}c@{}@{}c@{}@{}c@{}@{}c@{}@{}c@{}@{}c@{}@{}c@{}@{}c@{}@{}c@{}}%
    \multirow{2}{*}{\rotatebox{90}{C=4}} & 
    \resizebox{\subfigwidth}{!}{\graphCommunity{img/community_4_56_edge.csv}{img/community_4_56_node.csv}} &
    \resizebox{\subfigwidth}{!}{\graphCommunity{img/community_4_57_edge.csv}{img/community_4_57_node.csv}} &
    \resizebox{\subfigwidth}{!}{\graphCommunity{img/community_4_58_edge.csv}{img/community_4_58_node.csv}} &
    \resizebox{\subfigwidth}{\subfigwidth}{\graphCommunity{img/community_4_59_edge.csv}{img/community_4_59_node.csv}} &
    \resizebox{\subfigwidth}{!}{\graphCommunity{img/community_4_61_edge.csv}{img/community_4_61_node.csv}} &
    \resizebox{\subfigwidth}{!}{\graphCommunity{img/community_4_62_edge.csv}{img/community_4_62_node.csv}} &
    \resizebox{\subfigwidth}{!}{\graphCommunity{img/community_4_63_edge.csv}{img/community_4_63_node.csv}} &
    \resizebox{\subfigwidth}{!}{\graphCommunity{img/community_4_64_edge.csv}{img/community_4_64_node.csv}} \\ 
    
    &
    \resizebox{\subfigwidth}{\subfigwidth}{\plotNodes{img/H_community_4_56_feature_norm.csv}} &
    \resizebox{\subfigwidth}{\subfigwidth}{\plotNodes{img/H_community_4_57_feature_norm.csv}} & 
    \resizebox{\subfigwidth}{\subfigwidth}{\plotNodes{img/H_community_4_58_feature_norm.csv}} &
    \resizebox{\subfigwidth}{\subfigwidth}{\plotNodes{img/H_community_4_59_feature_norm.csv}} &
    \resizebox{\subfigwidth}{!}{\plotNodes{img/H_community_4_61_feature_norm.csv}} & 
    \resizebox{\subfigwidth}{!}{\plotNodes{img/H_community_4_62_feature_norm.csv}} & 
    \resizebox{\subfigwidth}{!}{\plotNodes{img/H_community_4_63_feature_norm.csv}} &
    \resizebox{\subfigwidth}{!}{\plotNodes{img/H_community_4_64_feature_norm.csv}} \\%
    
    \multirow{2}{*}{\rotatebox{90}{C=2}} & 
    \resizebox{\subfigwidth}{0.12\textwidth}{\graphCommunity{img/community_2_93_edge.csv}{img/community_2_93_node.csv}} &
    \resizebox{\subfigwidth}{!}{\graphCommunity{img/community_2_84_edge.csv}{img/community_2_84_node.csv}} &
    \resizebox{\subfigwidth}{!}{\graphCommunity{img/community_2_85_edge.csv}{img/community_2_85_node.csv}} &
    \resizebox{\subfigwidth}{0.12\textwidth}{\graphCommunity{img/community_2_86_edge.csv}{img/community_2_86_node.csv}} &
    \resizebox{\subfigwidth}{!}{\graphCommunity{img/community_2_92_edge.csv}{img/community_2_92_node.csv}} &
    \resizebox{\subfigwidth}{0.12\textwidth}{\graphCommunity{img/community_2_95_edge.csv}{img/community_2_95_node.csv}} &
    \resizebox{\subfigwidth}{0.12\textwidth}{\graphCommunity{img/community_2_94_edge.csv}{img/community_2_94_node.csv}} &
    \resizebox{\subfigwidth}{0.12\textwidth}{\graphCommunity{img/community_2_83_edge.csv}{img/community_2_83_node.csv}} \\
    
    &
    \resizebox{\subfigwidth}{\subfigwidth}{\plotNodes{img/H_community_2_93_feature.csv}} &
    \resizebox{\subfigwidth}{\subfigwidth}{\plotNodes{img/H_community_2_84_feature.csv}} & 
    \resizebox{\subfigwidth}{\subfigwidth}{\plotNodes{img/H_community_2_85_feature.csv}} &
    \resizebox{\subfigwidth}{\subfigwidth}{\plotNodes{img/H_community_2_86_feature.csv}} &
    \resizebox{\subfigwidth}{!}{\plotNodes{img/H_community_2_92_feature.csv}} & 
    \resizebox{\subfigwidth}{!}{\plotNodes{img/H_community_2_95_feature.csv}} & 
    \resizebox{\subfigwidth}{!}{\plotNodes{img/H_community_2_94_feature.csv}} &
    \resizebox{\subfigwidth}{!}{\plotNodes{img/H_community_2_83_feature.csv}} \\%
  \end{tabular}%
  \caption{Results on Community dataset predictions for the proposed methods, and the learned latent space, used for build adjacency matrix in the prediction.  The blue edges represent false negatives (\ie, not predicted edges), red edges represent false positives (\ie, additional predicted edges), and black edges are correctly predicted ones.  The graphs were normalized (\wrt scale and translation) for better visualization.}
  \label{fig:pred_community}%
\end{figure*}

\section{Prediction of Geometric Image}
\label{sec:pred_geo}
Finally, in Fig.~\ref{fig:pred_geometric_img}, we present an application, even fundamental, on segmentation where each of the connected components represents different objects. 
For this, we apply our GLN model on Geometric Image dataset, using size image of $20 \times 20$.
Besides, the white edges represent correct predictions, and light blue dashed edges are false negatives (\ie, not predicted edges).

\setlength{\subfigwidth}{0.35\linewidth}
\begin{figure*}
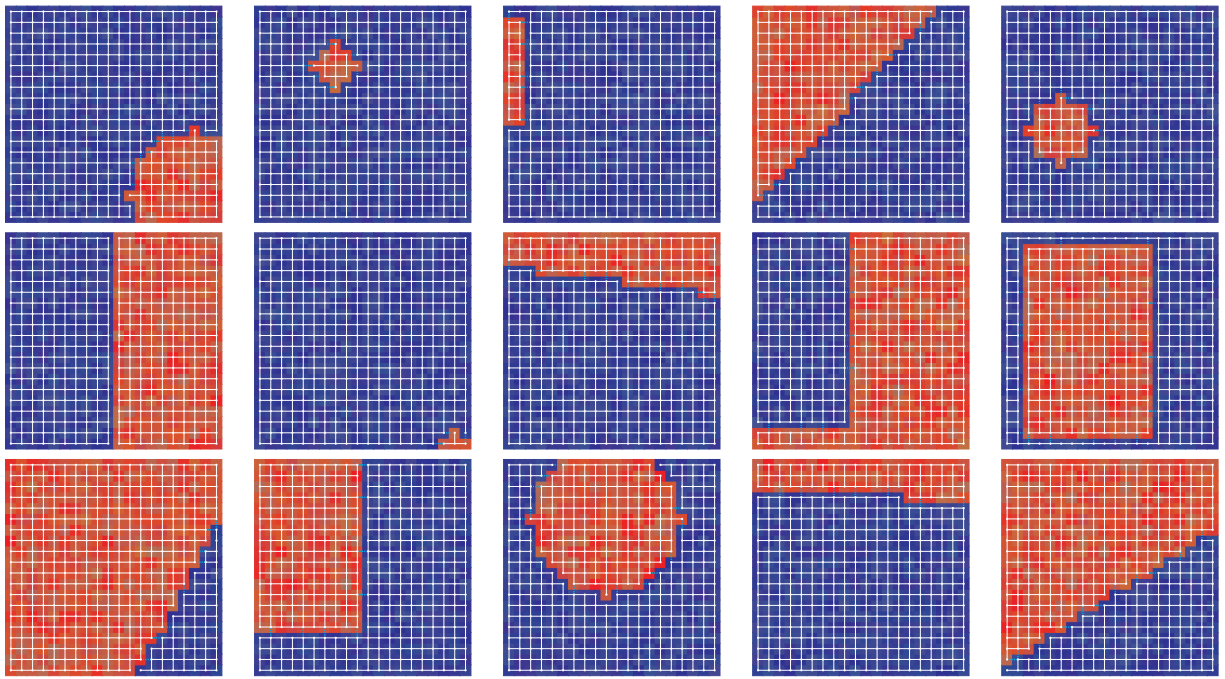

  \centering%
  \begin{tabular}{@{}cccccc@{}}
    & 
    \resizebox{\subfigwidth}{!}{\graphImg{img/img_feature_0.csv}{img/img_graph_edge_0.csv}} &
    \resizebox{\subfigwidth}{!}{\graphImg{img/img_feature_21.csv}{img/img_graph_edge_21.csv}} &
    \resizebox{\subfigwidth}{!}{\graphImg{img/img_feature_2.csv}{img/img_graph_edge_2.csv}} & 
    \resizebox{\subfigwidth}{!}{\graphImg{img/img_feature_4.csv}{img/img_graph_edge_4.csv}} &
    \resizebox{\subfigwidth}{!}{\graphImg{img/img_feature_7.csv}{img/img_graph_edge_7.csv}} \\
    &  
    \resizebox{\subfigwidth}{!}{\graphImg{img/img_feature_6.csv}{img/img_graph_edge_6.csv}} &  
    \resizebox{\subfigwidth}{!}{\graphImg{img/img_feature_10.csv}{img/img_graph_edge_10.csv}} &  
    \resizebox{\subfigwidth}{!}{\graphImg{img/img_feature_26.csv}{img/img_graph_edge_26.csv}} &  
    \resizebox{\subfigwidth}{!}{\graphImg{img/img_feature_30.csv}{img/img_graph_edge_30.csv}} &  
    \resizebox{\subfigwidth}{!}{\graphImg{img/img_feature_33.csv}{img/img_graph_edge_33.csv}} \\
    &  
    \resizebox{\subfigwidth}{!}{\graphImg{img/img_feature_34.csv}{img/img_graph_edge_34.csv}} &  
    \resizebox{\subfigwidth}{!}{\graphImg{img/img_feature_39.csv}{img/img_graph_edge_39.csv}} &  
    \resizebox{\subfigwidth}{!}{\graphImg{img/img_feature_44.csv}{img/img_graph_edge_44.csv}} &  
    \resizebox{\subfigwidth}{!}{\graphImg{img/img_feature_47.csv}{img/img_graph_edge_47.csv}} &  
    \resizebox{\subfigwidth}{!}{\graphImg{img/img_feature_50.csv}{img/img_graph_edge_50.csv}} \\
  \end{tabular}
  \caption{Predicted graphs using GLN on images with geometric shape of $20\times20$ pixels.  The image behind the graph corresponds to the input values at each node (RGB values), the white edges represent correct predictions, yellow dashed edges are false negatives (\ie, not predicted edges), and light blue dashed edges are false positives (\ie, additional predicted edges).}
  \label{fig:pred_geometric_img}
\end{figure*}

\end{document}